\documentclass[11pt]{article}
\usepackage{acl}
\usepackage{times}
\usepackage{latexsym}
\usepackage[T1]{fontenc}
\usepackage[utf8]{inputenc}
\usepackage{microtype}
\usepackage{inconsolata}
\usepackage{graphicx}
\usepackage{booktabs}
\usepackage{multirow}
\usepackage{array}
\usepackage{amsmath}
\usepackage{amssymb}
\usepackage{xspace}
\usepackage{float}
\usepackage{url}
\newcommand{\vp}{VideoPoints platform\xspace}
\newcommand{\embeddingmodel}{all-MiniLM-L6-v2\xspace}
\newcommand{\modelname}{Gemini-2.5 flash-lite\xspace}

  {\begin{itemize}%
    \setlength{\itemsep}{0pt}%
    \setlength{\parskip}{0pt}}%
  {\end{itemize}}

\newenvironment{description*}%
  {\begin{description}%
    \setlength{\itemsep}{0pt}%
    \setlength{\parskip}{0pt}}%
  {\end{description}}

  {\begin{enumerate}%
    \setlength{\itemsep}{0pt}%
    \setlength{\parskip}{0pt}}%
  {\end{enumerate}}

\title{Cite or Decline: A Strict Course-Grounded Chatbot for STEM Lecture Videos}

\author{S M Masrur Ahmed \\
  University of Houston\\ %/ Address line 1 \\
  \texttt{sahmed75@uh.edu} \\\And
  Jaspal Subhlok \\
  University of Houston \\ %/ Address line 1 \\
  \texttt{jaspal@uh.edu} \\}

\begin{document}
\maketitle
\begin{abstract}

Recorded lecture videos, often enhanced with search and summarization features, are a standard study resource. However, students cannot easily ask course specific questions or verify answers against an instructor's lecture. We report a semester-long deployment of \vp with a retrieval-augmented chatbot that answers from course lecture materials and returns timestamped citations. The chatbot retrieves only from the active course, uses chapter summaries to guide transcript ranking, and returns clickable timestamped citations. Students used it for quick lookups and exam review. Across 833 messages, 70.5\% included citations, none crossed a course boundary, and when no lecture evidence matched, the chatbot usually declined rather than answering. Among the users, citations were the most consistently useful feature, while practice-question generation was the strongest unmet request. We also evaluated the design on the real-world test split of EduVidQA, a public multimodal benchmark for lecture-video question answering. Our design improved correct-lecture retrieval by 6.3 percentage points over dense-only retrieval. Together, the results show that effective deployment depends on course isolation, supported citations, and alignment with students' study practices.

\end{abstract}

\section{Introduction}

Recorded lecture videos are an important resource in higher education that is employed to review difficult material and prepare for exams.
Prior work shows that indexed and searchable lecture videos can support STEM learning as students can jump to relevant segments of interest rather than replay full recordings \citep{barker2014,tuna2017ics}. \vp has supported this kind of intuitive navigation for over a decade through topic-based segmentation, captioning and search \citep{tuna2015,rahman2024}.
 
The state of the art still has a significant gap in achieving the broader goal of converting a collection of lecture videos into an interactive learning companion. A chapter index can show locations where a topic is discussed  but cannot answer a student's question. A general chatbot can answer student questions but the content may not match an instructor's framing. Furthermore, a chatbot cannot answer context specific logistical questions like the date and the style of a quiz.
 
The project began with a commitment to the instructors: answers had to come from the active course's lecture materials, and every answer had to point back to the relevant video segments. As a result, course grounding was a requirement, not just a technical preference. 

The approach to implementing and deploying the lecture-video chatbot in \vp was as follows. In Phase 1(Fall'25), \modelname was used to divide lecture videos into chapters and GPT-4.1 Mini to generate a summary for each chapter. These summaries were accessible to the students. In Phase 2 (Spring'26), the chapter summaries became an aid for ranking in the retrieval layer to generate course-grounded answers that best matched student questions. Chapter summaries thus serve as both student-facing navigation aids and retrieval anchors.

We organize the study around three research questions: \textbf{RQ1}, how students use a course-grounded video chatbot in real coursework; \textbf{RQ2}, when the chatbot provides citation-supported, course-isolated answers, when it refuses or fails to meet student requests, and which retrieval design choices those outcomes depend on; and \textbf{RQ3}, how students perceive chapter summaries, timestamped citations, trust, and exam usefulness, and what unmet study needs they report.

To answer these questions, we combine production logs, retrieval traces, survey responses, and a diagnostic audit of citation support and refusal behavior. We also evaluate the retrieval design on EduVidQA, a public lecture-video question-answering benchmark \citep{ray2025eduvidqa}. This controlled evaluation compares the full system with retrieval baselines under the same corpus, questions, and retrieval budget.

The paper makes three contributions. \textbf{First}, it reports a semester-long, multi-course deployment and characterizes how students used and perceived a citation-first lecture-video chatbot. \textbf{Second}, it evaluates the system's grounding and its retrieval design separately. A production audit estimates how often the cited lecture segments support the generated answer, while a controlled benchmark compares the retrieval design with dense-only and course-unrestricted baselines. \textbf{Third}, it identifies transferable deployment lessons about course isolation, question reformulation, and the mismatch between retrieval-based question answering and students' task-oriented requests. The contribution is not a new RAG architecture. It is evidence about how a known architecture behaves under real instructional and deployment constraints.

\section{Related Work}

\paragraph{Lecture video navigation.}
Lecture video systems have long focused on helping students find relevant content of interest. \vp introduced topic-based segmentation and indexed lecture navigation for STEM coursework \citep{tuna2015,tuna2017ics}. Later work added AI-generated visual and textual summaries to improve navigation and review \citep{rahman2020,rahman2024}. Other systems use visual anchors, structural cues, or transcript search to support non-linear access to educational videos \citep{yadav2016,das2019}. These systems improve access to lecture content, while the goal of the work presented in this paper is a course-grounded conversational chatbot deployed over lecture videos. LLM-generated summaries can support student review. Prior work found that students who received both lecture videos and AI-generated summaries reported favorable perceptions and, in some settings, improved learning outcomes \citep{gonzalez2023}.

\paragraph{Retrieval-augmented generation in education.}
Retrieval-augmented generation provides a natural method for answering questions from external materials \citep{lewis2020}. Educational RAG systems can improve course specificity, but they also raise concerns about trust, citation quality, and mismatch between generated answers and instructional context. \citet{tanner2024} study retrieval-augmented question answering over lecture videos in an evaluation setting. SyllabusQA studied question answering over course logistics documents \citep{fernandez2024}. Educational chatbots have supported student support, tutoring, and course assistance \citep{Taneja2024,roca2024,rouhani2025}.
Our study complements these efforts by focusing on production use: real courses, real student questions, timestamped video citations, and instructor-driven course isolation.

\section{System and Deployment Setting}

\subsection{\vp}
The deployment used VideoPoints, a production lecture-video platform in STEM courses at a large public university. The platform provides thumbnails, captions, search, transcripts, keywords, visual highlights, as well as instructor tools for adjusting chapter boundaries~\citep{rahman2024,biswas2025visual}. It also segments lectures into topic-based chapters using \modelname. The deployment discussed in this paper leveraged this existing VideoPoints infrastructure. A complete view of the chatbot interface is provided in Appendix~\ref{app:interface} (Figure~\ref{fig:interface}).

\subsection{Chapter Summaries}

Each lecture chapter received an LLM-generated title and summary using GPT-4.1 Mini. Summaries were displayed on the platform player page after hovering over the chapter index. The practical goal was simple: students should be able to decide whether a chapter is relevant before watching it. This design was tested in Phase 1 before chatbot deployment. An example chapter summary view of the interface is provided in
Appendix~\ref{app:interface} (Figure~\ref{fig:sum_interface}).

The summaries served a second role in the Spring chatbot. New summaries were generated for the Spring 2026 lectures using the same prompt evaluated in Phase 1. Because each summary was generated once and stored with its chapter, the chatbot could use it as a relevance signal without generating new summaries at query time. Chapter summaries provide shorter and cleaner descriptions of lecture topics than raw transcript spans. The system therefore uses query-summary similarity as a soft prior when ranking transcript and slide-text chunks. This prior does not exclude any material from the active course. It only raises the scores of chunks from chapters whose summaries better match the student's question.

\subsection{Citation-First RAG Pipeline}\label{sec:pipeline}

The design of lecture video support chatbot reflects the following key deployment constraints. First, instructors required course-only answers. Second, citations had to be clickable and tied to timestamps. Third, student text had to be handled without retaining identifying information. Fourth, the system had to respond fast enough for routine study use. These constraints ruled out a general chatbot and motivated a retrieval-only design.

Figure~\ref{fig:pipeline} summarizes the deployed pipeline. First, the student
question is encoded using all-MiniLM-L6-v2 embeddings \citep{reimers2019}.
Second, retrieval is restricted to the active course. Third, chapter summaries
are used as a soft relevance prior rather than a hard filter: all transcript and
slide-text chunks remain eligible, but chunks from chapters with better-matching
summary bullets are ranked higher. The final score combines dense similarity,
BM25 lexical similarity, and the chapter-summary prior. This design preserves
recall while using summaries as cleaner retrieval anchors for noisy lecture
transcripts. Fourth, the retrieved context is passed to \modelname for answer generation, with the prompt requiring timestamped grounding. Finally, the system returns an answer with timestamped citations that link back to the relevant chapter in the lecture video.

\begin{figure}[t]
\centering
\fbox{
\begin{minipage}{0.92\linewidth}
Student question $\rightarrow$ course filter $\rightarrow$ chapter-summary prior
$\rightarrow$ dense/BM25/prior ranking $\rightarrow$ top-$k$
transcript/slide chunks $\rightarrow$ answer generation $\rightarrow$
timestamped citations.
\end{minipage}
}
\caption{Cite or Decline chatbot retrieval pipeline. Appendix~\ref{sec:design} gives the scoring function and model choices.}
\label{fig:pipeline}
\end{figure}

\subsection{Deployment Phases and Usage}

Table~\ref{tab:deployment} summarizes the deployment and student usage. Phase 1 ran in Fall 2025 and collected summary feedback from 128 respondents. The chatbot was not yet deployed. Phase 2 ran in Spring 2026 and collected chatbot logs over 93 days. It also collected survey responses from 41 students, including 22 self-reported chatbot users. Both phases ran on our lab's existing infrastructure, so the only marginal cost was model API usage: approximately \$100 in total, including development. 

\begin{table}[t]
\centering
\small
\begin{tabular}{lcc}
\toprule
Metric & Phase 1 & Phase 2 \\
\midrule
Focus & summaries & chatbot \\
Term & Fall 2025 & Spring 2026 \\
Survey respondents & 128 & 41 \\
Confirmed chatbot users & -- & 22 \\
Sessions & -- & 680 \\
Active sessions & -- & 233 \\
Messages & -- & 833 \\
Active courses & 10 & 8 \\
Videos / chapters & 109/645 & 96/570 \\
\bottomrule
\end{tabular}
\caption{Deployment overview. Phase 1 measured perception of chapter summaries before chatbot release. Phase 2 measured chatbot usage, retrieval traces, and student perception.}
\label{tab:deployment}
\end{table}

\section{Evaluation Methodology}

The evaluation combines survey data, system logs, retrieval traces, de-identified case analysis, a diagnostic audit of production messages, and a controlled offline retrieval evaluation. It does not measure learning outcomes or include a randomized control group. Therefore, our claims focus on deployment behavior, perceived usefulness, grounding coverage, and production failure modes.

\paragraph{Survey measures.}
The Fall 2025 survey measured summary accuracy, navigation value, review value, desired summary length, and pre-deployment interest in a chatbot, while the Spring 2026 survey measured chatbot experience, including ease of use, trust, citation usefulness, exam usefulness, answer-length preference, textbook integration interest, and practice-question interest. All survey items used a 1-5 Likert scale. We report the number of responses per item, as some respondents did not answer every item. We asked instructors to email their students about the features and to circulate the survey links, and a banner on the VideoPoints website also linked to them. Participation was strictly optional.

\paragraph{Privacy and Log measures.}
To protect student privacy, session identifiers were salted and hashed before analysis. Raw logs remained on local infrastructure. For the diagnostic audit, questions and responses were de-identified and screened for direct identifiers before the screened text was sent to an external model API. We report no verbatim student examples in this paper, and all case descriptions are paraphrased and de-identified. Each stored chat record contains the student's question, the chatbot's response, and any returned video matches.

\paragraph{Diagnostic audit.}
Citation presence in the production logs does not show whether the cited lecture segments support the answer. We therefore conducted a stratified LLM-assisted audit of 224 unique messages. The sample included all four citation-refusal outcome groups, with the two smaller groups and all imperative requests audited in full. The audit evaluated citation support, evidence relevance, refusal appropriateness, and implicit refusals missed by the original regex. Sampling weights mapped the audited groups back to the 833-message population. These judgments provide diagnostic estimates rather than human-validated factuality labels. Appendix~\ref{sec:audit} reports the sampling design, judge configuration, and confidence-interval procedure.

\paragraph{Offline retrieval evaluation.}
Production traffic could not provide a live baseline without course isolation because instructors required course-only retrieval. We therefore evaluated the retrieval design on the real-world split of EduVidQA \citep{ray2025eduvidqa}. The evaluation contains 269 questions from 99 lectures, spanning seven of the ten courses, within a corpus of 139 videos, 10 courses, 1,062 chapters, and 5,924 chunks. We used an LLM to group the videos into courses and to generate chapter boundaries and summaries; these assignments were produced once and reused unchanged across all retrieval arms. Every retrieval arm used the same corpus, questions, embeddings, chunking, context budget, and retrieval budget. This experiment reconstructs the deployed retrieval design. Appendix~\ref{sec:ablation_details} gives the corpus, arms, and metric definitions.

\section{Results}
\label{sec:results}

We introduce two terms relevant to the chatbot outcomes.  A \emph{no-citation}  implies that a retrieval returned no relevant citation. A \emph{refusal} explicitly states that the available course evidence is insufficient based on pattern matching over the response text. The two are separate outcomes: some responses carried citations and still refused, and some carried no citation without an explicit refusal.

\subsection{RQ1: Usage and Study Behavior}
\label{sec:rq1}

Table~\ref{tab:deployment_activity} summarizes Phase 2 usage. Three observations matter. \textit{(i)} Of the 680 sessions, 447 contained no stored exchange and should not be interpreted as active use. \textit{(ii)} Usage was concentrated, with the most active course producing 78.8\% of all messages. \textit{(iii)} Students clicked timestamped citations 313 times, although the logs do not show how much of the cited videos they watched. These results illustrate why empty sessions, course concentration, and citation clicks should be reported separately from total traffic.

\begin{table}[ht]
\centering
\small
\begin{tabular}{lr}
\toprule
Deployment measure & Value \\
\midrule
Sessions & 680 \\
\quad With at least one message & 233 \\
\quad With no stored exchange & 447 \\
Messages & 833 \\
Messages from most active course & 656 (78.8\%) \\
\midrule
Active session depth: mean / median / max. & 3.58 / 2 / 68 \\
Sessions with one message & 108 \\
Sessions with at least five messages & 38 \\
Citation clicks & 313 (37.6\%) \\
\bottomrule
\end{tabular}
\caption{Phase 2 deployment activity from February 23 to May 26, 2026, with recorded activity on 60 days. The click figure is clicks per message.}
\label{tab:deployment_activity}
\end{table}

Usage also aligned with assessment periods. A five-day window from March 23 to March 27 contained 328 messages, or 39.4\% of all messages, and fell within the university's midterm dates of March 23--31.
A second high-use period from May 4 to May 11 overlapped finals dates of May 6--12. The longest session contained 68 turns in which a student reviewed an exam outline topic by topic. These patterns show that students used the chatbot for sustained exam preparation as well as short information requests. However, the temporal alignment does not show that exams caused the increase in use or that chatbot use improved learning.

Table~\ref{tab:interaction_form} shows how message form related to system outcomes. Questions and keyword fragments together accounted for 688 of 833 messages, or 82.6\%, and had similar refusal and no-citation rates. Imperative messages were different. For example, a request to generate a practice quiz is imperative because it asks the chatbot to perform a task rather than answer a course-content question. Imperative messages had higher odds of refusal than non-imperative messages (odds ratio 2.74, 95\% CI 1.55--4.86, Fisher's exact $p = 0.0007$). They also had higher odds of returning no citation (odds ratio 3.74, 95\% CI 2.10--6.68, $p < 10^{-5}$). In practical terms, the chatbot handled direct questions and keyword searches more successfully than commands asking it to perform a task.

\begin{table}[ht]
\centering
\small
\begin{tabular}{lrrr}
\toprule
Interaction form & $n$ & Refusal & No citation \\
\midrule
Question & 469 & 27.5\% & 27.7\% \\
Keyword fragment & 219 & 25.6\% & 26.9\% \\
Statement/other & 94 & 31.9\% & 28.7\% \\
Imperative command & 51 & 51.0\% & 58.8\% \\
\bottomrule
\end{tabular}
\caption{Interaction form and system outcomes. Imperative messages produced more refusals and no-citation outcomes than non-imperative messages. Interaction-form labels agreed with an independent automated labeler at $\kappa = 0.876$.}
\label{tab:interaction_form}
\end{table}

Intent labels indicate what students asked about, most often definition recall, follow-up clarification, and exam preparation. Because agreement with an independent automated labeler was low ($\kappa = 0.275$), these categories are exploratory and no main finding relies on them; Appendix~\ref{sec:labels} discusses the categories and their reliability.

Together with the de-identified case analysis, the interaction patterns suggest three ways students approached the chatbot. Some treated it as a search box and entered keyword fragments. Others treated it as a question-answering chatbot and requested explanations or comparisons. A third group expected an agent that could generate, grade, plan, remember earlier turns, or operate over ranges of lectures. The system served the first two patterns more successfully than the third.

\subsection{RQ2: Citation Support, Refusal, and Retrieval Design}
\label{sec:rq2}

RQ2 addresses three connected questions: \textit{(i)} whether the chatbot remained within the active course, \textit{(ii)} whether its citations supported the answers and its refusals were appropriate, and \textit{(iii)} which retrieval choices contributed to these outcomes. We answer the first question from production logs, the second through the diagnostic audit described in the methodology, and the third through the controlled EduVidQA evaluation.

\paragraph{Course isolation and citation coverage.}
The chatbot returned citations for 587 of 833 messages, giving a citation coverage of 70.5\% (95\% CI 67.3--73.5). Each cited message displayed exactly seven citations, a fixed output cutoff, producing 4,109 citation events. None pointed outside the active course. This verifies course isolation in production, but it does not show whether the cited segments supported the answers.

Citation presence and refusal were separate outcomes. The logs contained 548 cited responses without a detected refusal, 39 cited refusals, 44 uncited responses without a detected refusal, and 202 uncited refusals. These groups use regex-based refusal labels and are not correctness categories. Table~\ref{tab:grounding_results} therefore separates production measures from the audit estimates.

\begin{table}[ht]
\centering
\small
\setlength{\tabcolsep}{3pt}
\begin{tabular}{@{}p{0.52\columnwidth}rr@{}}
\toprule
Measure & Estimate & 95\% CI \\
\midrule
\multicolumn{3}{l}{\textit{Production logs}} \\
Citation coverage & 70.5\% & 67.3--73.5 \\
Cross-course citations & 0/4,109 & -- \\
Explicit refusal when uncited & 82.1\% & 76.8--86.4 \\
\midrule
\multicolumn{3}{l}{\textit{Diagnostic audit}} \\
Fully supported & 65.0\% & 55.0--75.0 \\
Fully or partially supported & 86.3\% & 78.8--93.8 \\
Unsupported by citations & 11.3\% & 5.0--18.8 \\
Appropriate refusal & 61.9\% & 51.4--71.8 \\
Implicit refusal among uncited
non-refusals & 61.4\% & 47.7--75.0 \\
\bottomrule
\end{tabular}
\caption{Production measures and weighted diagnostic audit estimates. Citation-support estimates apply to cited responses without a detected refusal. Log intervals are Wilson; audit intervals are stratified bootstrap percentile intervals. The audit used automated judgments, not human annotations.}
\label{tab:grounding_results}
\end{table}

\paragraph{Citation support and refusal quality.}
The audit estimated that 65.0\% of cited non-refusal answers were fully supported and 86.3\% were fully or partially supported. An estimated 11.3\% were unsupported, and the small remainder could not be judged. The chatbot stayed within the active course, but a citation did not always support the generated answer.

The same distinction applies to refusals. Of the 246 uncited messages, 202 contained an explicit refusal detected by the original regex. However, 27 of the remaining 44 contained implicit refusal language that the regex missed. The audit also estimated that 61.9\% of refusals were appropriate, with the other cases judged unclear or inappropriate. The system often declined when citations were absent, but refusal frequency alone did not establish that each refusal was appropriate.

\paragraph{Failure, recovery, and task expectations.}
Of the 170 no-citation turns followed by another turn, 93 were followed by a citation-present response. This is consistent with question reformulation, although the logs cannot establish whether each following message addressed the same request. Case analysis also identified short or misspelled questions, broad multi-chapter requests, generation requests, and requests for conversational memory as recurring difficulties. Case analysis confirmed at least one cited response that was incorrect because the underlying lecture evidence was noisy. Together, these cases show that both successful retrieval and refusal require quality checks.

Task-oriented requests exposed a further boundary. At least 61 messages, or 7.3\%, expressed an agentic expectation such as generating practice questions, grading answers, or remembering previous turns. These messages had a 78.7\% refusal rate and an 86.9\% no-citation rate, compared with 25.0\% for both outcomes among other messages. Because agreement for the agentic label was moderate ($\kappa=0.590$), these comparisons are descriptive. Even so, they show that retrieval-based question answering did not match all student expectations.

\subsubsection{Controlled Offline Retrieval Evaluation}
\label{sec:ablation}

The final part of RQ2 examines which retrieval choices mattered. Table~\ref{tab:eduvidqa_ablation} compares the full design with controlled alternatives on EduVidQA. All systems used the same data, embeddings, chunking, and retrieval budget, as described in the methodology.

\begin{table}[ht]
\centering
\small
\setlength{\tabcolsep}{3pt}
\begin{tabular}{lrrrrr}
\toprule
System & TS hit & Vid hit & P@5 & MRR & nDCG@5 \\
\midrule
Dense only & 0.402 & 0.684 & 0.103 & 0.250 & 0.283 \\
+ BM25 & 0.442 & 0.710 & 0.103 & 0.287 & 0.311 \\
\textbf{+ Summary prior} & \textbf{0.457} & \textbf{0.747} &
\textbf{0.129} & \textbf{0.293} & \textbf{0.330} \\
No course isolation & 0.338 & 0.569 & 0.095 & 0.209 & 0.239 \\
\bottomrule
\end{tabular}
\caption{EduVidQA retrieval results on the real-world split ($n=269$, $k=5$).
TS and Vid denote timestamp and source-video hit; P@5 is the fraction of the
five retrieved chunks that come from the source video.}
\label{tab:eduvidqa_ablation}
\end{table}

The full design outperformed dense-only retrieval on timestamp hit, video hit, MRR, and nDCG@5, with paired-test $p$-values from 0.010 to 0.032. The effect sizes were small ($d_z=0.14$ for MRR and $d_z=0.16$ for nDCG@5), so this result supports a modest improvement rather than a decisive one. Adding BM25 produced most of the timestamp-level gain. The summary prior produced a further descriptive increase across all reported metrics. Because we do not report a paired significance test for the BM25-versus-full comparison, we treat the prior as a ranking refinement rather than making a statistical claim about its individual contribution.

Course isolation produced the clearest result. Removing it reduced video hit from 0.747 to 0.569, a decrease of 17.8 percentage points. This was the largest decrease among the tested changes and was statistically supported for video hit ($p\leq0.0005$) and timestamp hit ($p\leq0.034$). The unrestricted system still retrieved above-threshold evidence for 98.9\% of questions, but that evidence often came from the wrong course, so confidence-style metrics can reward exactly the behavior instructors prohibited. Course isolation mattered more than either scoring refinement and directly supported the instructor's deployment requirement.

\subsection{RQ3: Student Perceptions and Unmet Study Needs}
\label{sec:rq3}

The survey analysis tests two hypotheses. \textbf{H1} predicts that students value chapter summaries for both navigation and review but prefer different summary lengths for the two tasks. \textbf{H2} predicts that students who used the chatbot rate timestamped citations, trust, and exam usefulness positively. Table~\ref{tab:survey_results} reports item means for both phases.

\begin{table}[t]
\centering
\small
\begin{tabular}{lrr}
\toprule
Survey item & Mean & Top-2-box \\
\midrule
\multicolumn{3}{l}{\textit{Phase 1: chapter summaries}} \\
Chapters help find content & 4.67 & 92.0\% \\
Perceived summary accuracy & 4.34 & 81.4\% \\
Summaries aid navigation & 4.43 & 85.0\% \\
Summaries aid review & 4.42 & 85.8\% \\
\midrule
\multicolumn{3}{l}{\textit{Phase 2: chatbot users}} \\
Easy to use & 4.64 & 95.5\% \\
Trust in accuracy & 4.27 & 90.9\% \\
Citation links helpful & 4.55 & 100.0\% \\
Helpful for exams/quizzes & 4.55 & 86.4\% \\
Want practice questions + grading & 4.59 & 86.4\% \\
\bottomrule
\end{tabular}
\caption{Student perception results on a 1(Highly Disagree)--5(Highly Agree) scale. Phase 1 results use item-level $n=112$--113 of 128 respondents. Phase 2 results use $n=22$.}
\label{tab:survey_results}
\end{table}

Fall results support \textbf{H1}. For navigation, 75 of 112 respondents preferred the existing length of about five sentences. For review, only 48 of 113 preferred the existing length, while 61 wanted longer or much longer summaries. The same artifact is therefore about right for one task and too short for the other, which argues for task-dependent summary length.

Spring results support \textbf{H2}. Among the 22 self-reported chatbot users, ease of use received a mean of 4.64/5, citation usefulness and exam usefulness both 4.55/5, and trust in accuracy 4.27/5. Citation links were the most consistent item, with every respondent rating them 4 or 5, and trust was rated slightly below ease of use, though the difference is small relative to the sample size.

The strongest unmet need was practice-question generation and grading, rated 4.59/5 among 22 respondents. This aligns with the production logs, where task-oriented requests frequently asked the chatbot to generate practice exams, quiz students, or evaluate answers, and where those requests were most often refused. Answer-length preferences were divided, with no option receiving a majority among the 22 users. Together, these findings indicate demand for instructor-controlled practice activities and configurable response length.

These findings should be interpreted within the evaluation design. The surveys measure self-reported perceptions rather than controlled learning outcomes, the Spring results represent only 22 self-reported users, and survey responses could not be linked to individual usage logs. Even so, the survey and production results point to the same practical distinction: students valued grounded citations, but they also expected study support beyond course-grounded question answering.

\section{Discussion}

The main design lesson concerns retrieval scope. Instructors required the chatbot to answer only from the active course, and the production system met this requirement. The controlled evaluation also showed that removing course isolation degraded retrieval more than changing either scoring component. At the same time, the unrestricted system often retrieved high-confidence evidence from the wrong course. Retrieval confidence should therefore not be treated as evidence of correct course grounding.

Chapter summaries provided a useful but limited structural signal. Lecture transcripts are often noisy and conversational, while summaries give a cleaner description of each chapter's content. The summary prior produced small and consistent retrieval improvements, but its individual contribution is reported descriptively. It should therefore be treated as a ranking refinement rather than the main source of retrieval quality.

The evaluation also shows that citation coverage is different from citation support. A response can contain citations but the cited lecture video sections may not fully support the content of the response.
%without being fully supported by them.
Similarly, frequent refusal does not mean that every refusal is appropriate. These properties should be evaluated separately, and the current automated audit should be followed by human evaluation.

Student behavior exposed a further boundary. Students used the chatbot for searches, course questions, and exam review. However, some students also expected the chatbot to generate practice materials, grade answers, plan study, and remember past interactions. The retrieval-based design served direct questions successfully but was not designed for task-oriented requests.
%so the feature students wanted most was the one it could not provide. 
Educational chatbots therefore need clear boundaries and instructor-controlled tools for tasks beyond grounded question answering.

\section{Conclusion}

This paper presented a two-phase deployment of chapter summaries and a course-isolated lecture-video chatbot in STEM courses. Production logs showed both short information requests and sustained exam-review sessions, while self-reported users rated timestamped citations positively. All returned citation events stayed within the active course, but 246 of 833 messages received no citation, and the diagnostic audit found that some cited answers were not fully supported. These findings show that course isolation, citation presence, citation support, and refusal appropriateness should be evaluated separately. The study measures usage and perception rather than learning gains, but it provides practical evidence for designing citation-first educational chatbots and for identifying study tasks that require instructor-controlled support beyond question answering.

\section*{Limitations}

This study has several limitations. First, it was conducted at one institution, and active use was concentrated: one large introductory course produced 656 of 833 messages, so the results should not be treated as uniform across departments or instructors. Second, the unit of analysis is the session rather than the student. No user identifiers or enrollment denominators were available, so unique-user counts, repeat-use curves, and adoption rates cannot be computed.

Third, the survey results measure perception, not learning outcomes, and we cannot claim that the chatbot improved exam performance. Fourth, the interaction-form, intent, refusal, and agentic labels were produced by deterministic ordered rules over the query text together with regex-based refusal detection. Their agreement with an independent automated labeler was high for interaction form ($\kappa = 0.876$) and refusal ($\kappa = 0.806$), moderate for agentic expectation ($\kappa = 0.590$), and low for primary intent ($\kappa = 0.275$). We therefore report intent labels as exploratory descriptive patterns rather than validated categories, and the comparison measures consistency between two automated methods rather than accuracy against human annotation.

Fifth, citation generation does not prove citation use, factual correctness, or video watching. Sixth, the citation-support and refusal-appropriateness estimates come from an LLM-assisted diagnostic audit rather than human annotation, so we still do not report a human-validated factuality rate. The logs include at least one confirmed grounded-but-incorrect answer, which shows that citations alone do not guarantee correctness.

Seventh, the offline retrieval evaluation is a controlled reconstruction on a public benchmark rather than a replay of production traffic. It supports comparisons between design choices but does not reproduce the production courses, questions, or retrieval environment. Eighth, response latency was not instrumented during the deployment window, so the latency requirement is reported as a design constraint rather than a measured property. Finally, because deployment was voluntary, engaged students are likely overrepresented.

\section*{Ethical Considerations}

The system was deployed in a classroom setting, so privacy and academic integrity were central concerns. Chatbot logs were analyzed at the session level, and session identifiers were salted and hashed before export. The dataset contained no student identifiers, and survey responses were anonymous and not joinable to logs. The surveys were IRB-approved, participation was optional, and students consented before responding.

The diagnostic audit sent anonymized production questions and responses to an external model API for diagnostic labeling. An automated privacy scan ran before any request and found no direct identifiers, and raw session identifiers were not present in the export and were never transmitted. The audit artifacts retain hashed session keys, model identifiers, prompt hashes, and parse status. These safeguards reduce disclosure risk, but automated scanning cannot guarantee that free-form student text contains no sensitive information. No verbatim student text appears in this paper; all case descriptions are paraphrased and de-identified.

The deployment also surfaced assessment-related behavior. Some students pasted multiple-choice or true-false items into the chatbot. We report this as a deployment reality rather than as evidence of misconduct, since some items may come from review materials or ungraded practice. The behavior still has design implications. Educational RAG systems should include instructor controls, assessment-mode policies, and explain-without-answer options when deployed alongside graded work. 

AI was used as a writing assistant to improve grammar, sentence structure, clarity, and readability. The authors reviewed and verified all resulting text, factual claims, numerical results, and references and take full responsibility for the final manuscript.

\section*{Acknowledgment}
The authors express sincere gratitude to all current and former members of the Videopoints team, especially Jatindera Singh Walia and Dipayan Biswas.
We also acknowledge the encouragement and support of Videopoints faculty user participants, in particular Dr. Richard Knapp, Dr. Chad Wayne, Dr. Jokubas Ziburkus, and Dr. Pranav Mantini.
Partial support was received from the National Science Foundation under award NSF-SBIR-1820045. Partial support was also received in the form of a  University of Houston Teaching Innovation Program (TIP)  Grant. This work was completed in part with resources provided by the Research Computing Data Core at the University of Houston.

\bibliography{ref}

\newpage
\appendix
\section{Appendix}

\subsection{Additional Figures and Interface}
\label{app:interface}

\begin{figure}[ht]
    \centering
    \includegraphics[width=\columnwidth]{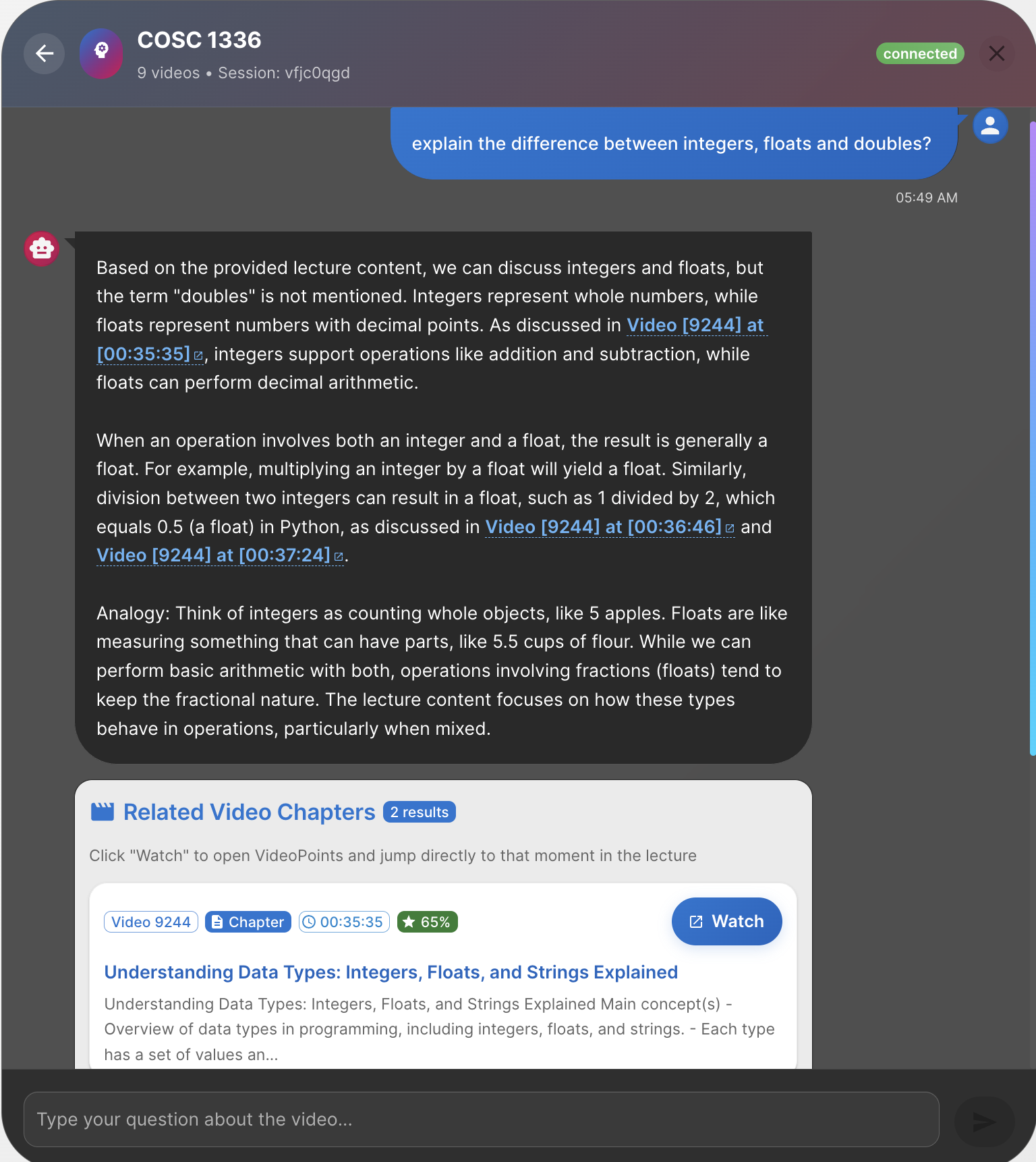}
    \caption{The chatbot interface used in the study with a test example.}
    \label{fig:interface}
\end{figure}

\begin{figure}[ht]
    \centering
    \includegraphics[width=\columnwidth]{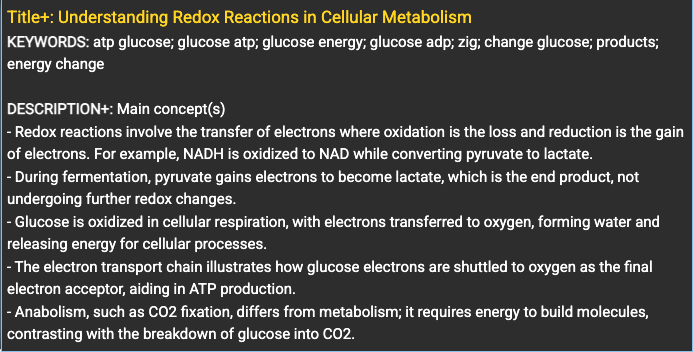}
    \caption{The chapter summary interface used in the study.}
    \label{fig:sum_interface}
\end{figure}

\begin{figure}[ht]
    \centering
    \includegraphics[width=\columnwidth]{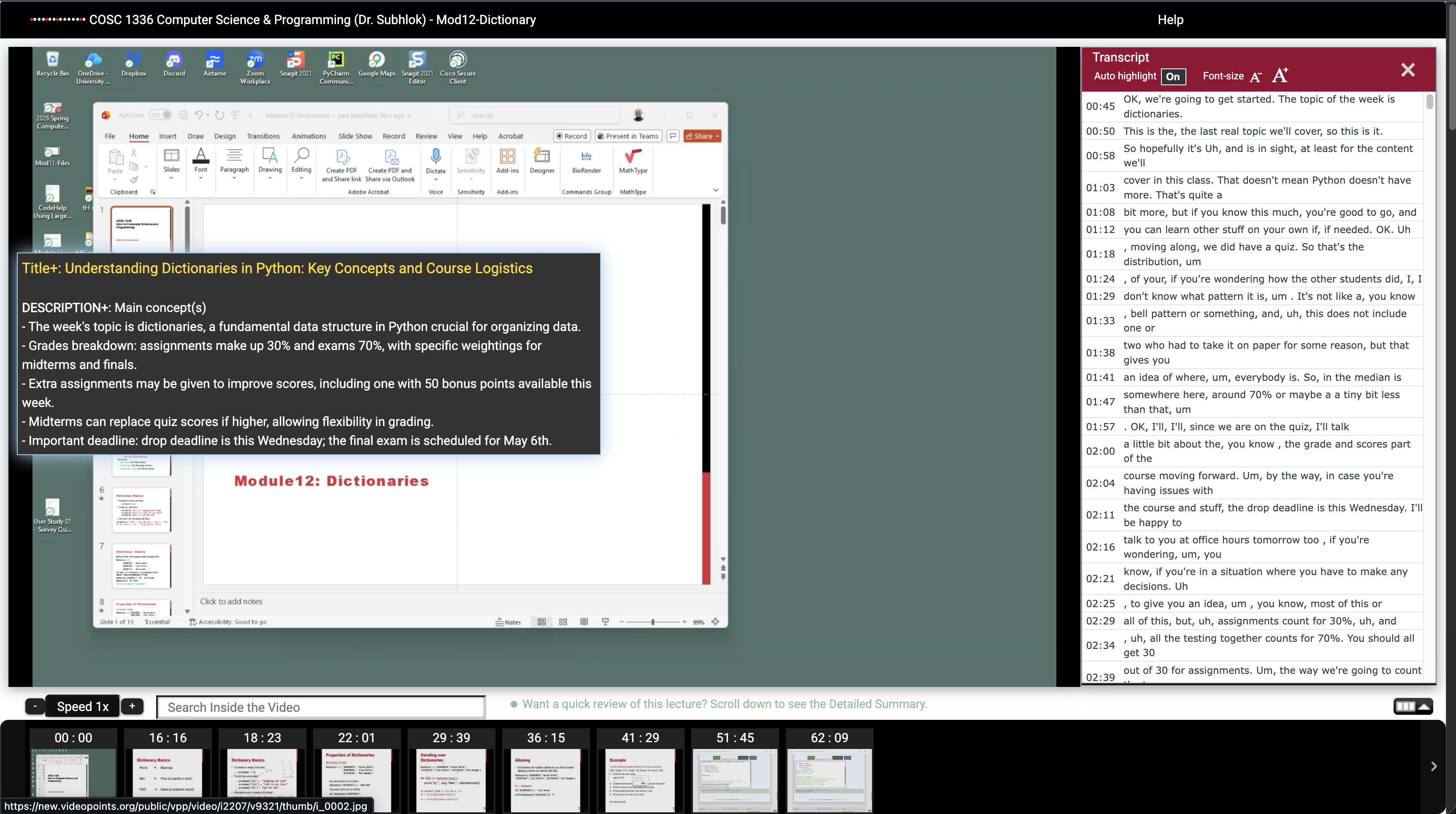}
    \caption{The \vp interface used in the study.}
    \label{fig:vp_interface}
\end{figure}

\begin{figure}[ht]
\centering
\includegraphics[width=\linewidth]{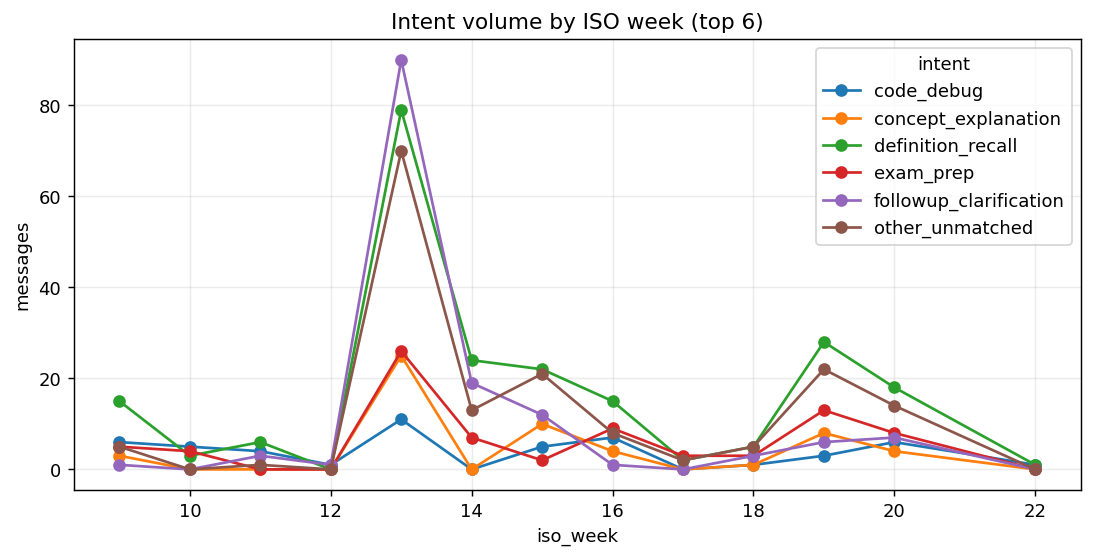}
\caption{Intent volume by ISO week for the top six exploratory intent
categories. The peak in week 13 contains the March 23--27 window, which held
328 messages and fell within the midterm dates (March 23--31). A second high-use period, May 4--11, overlapped the finals dates (May 6--12).}
\label{fig:intentweek}
\end{figure}

Figure~\ref{fig:interface} presents the chatbot interface used by students, and Figure~\ref{fig:sum_interface} presents the chapter-summary interface introduced in Phase 1, and Figure~\ref{fig:vp_interface} shows the surrounding platform interface. Figure~\ref{fig:intentweek} shows intent volume for the top six exploratory intent categories.

\subsection{Design Details}
\label{sec:design}

This appendix separates three evidence sources. Production logs establish
observable system behavior, the deployed codebase establishes model and
implementation choices, and the offline reconstruction supports the
comparisons in Section~\ref{sec:ablation} without being a replay of production
retrieval.

\subsubsection{Scoring Function}

Candidate chunks are scored with three per-query min--max normalized signals:
\begin{equation}
s(c \mid q)
=
w_d\,\widehat{d}(q,c)
+
w_b\,\widehat{b}(q,c)
+
w_p\,\widehat{p}\big(q,\operatorname{ch}(c)\big),
\label{eq:fusion}
\end{equation}
where $\widehat{d}$ is dense cosine similarity under \embeddingmodel
embeddings, $\widehat{b}$ is BM25 lexical relevance \citep{robertson2009}, and
$\widehat{p}$ is the chapter-summary prior of the chapter
$\operatorname{ch}(c)$ containing chunk $c$, with $w_d = 1.0$, $w_b = 1.25$,
and $w_p = 0.4$. Per-query normalization is necessary because BM25 scores are
unbounded while cosine similarities are not. Note that the lexical channel is
weighted above the dense channel.

The prior is computed at the bullet level: each summary bullet is embedded
separately, and a chapter's prior is the maximum query-to-bullet similarity.
Every chunk inherits the prior of its chapter. Two edge cases preserve the
candidate pool. A chapter whose summary fails validation is dropped from the
prior index while its chunks remain eligible (19 chapters), and a chunk whose
chapter has no usable summary inherits the per-query minimum prior rather than
zero, because cosine similarity can be negative and a zero fill would rank
such a chunk above a genuinely off-topic chapter after normalization (100
chunks). The candidate pool is identical before and after the prior is
computed.

\subsubsection{Model Choices and Instrumentation}

The embedding model is \embeddingmodel, selected because it runs locally
within the lab server's compute and memory limits and adds no per-query API
cost. The answer generator is \modelname, selected to keep generation cost
within budget; answer generation cost approximately \$55 and summary
generation approximately \$45, both including development use. Every
citation-present response displayed exactly seven citations, which is
consistent with a fixed output cutoff, and citations use the format
\texttt{[SOURCE n, HH:MM:SS]}. Response latency was not instrumented during
the deployment window, so the latency requirement in
Section~\ref{sec:pipeline} is a design constraint rather than a measured
property.

\subsection{Automated Labeling and Agreement}
\label{sec:labels}

Four message-level variables were used: interaction form (question, keyword
fragment, imperative command, or statement/other), refusal (binary, from regex
detection over the response text), agentic expectation (binary), and primary
intent (14 categories with a residual class). All were produced by
deterministic ordered rules over the lowercased query text and turn position;
no LLM produced them. The intent taxonomy was derived from a manual reading of
50 queries, applies rules in a fixed precedence order, and routes unmatched
messages to a residual category holding 19.3\% of messages, so it is
exhaustive by construction but not conceptually non-overlapping. Interaction
form and agentic expectation are separate axes: a request to generate a quiz
phrased as a question is a question in form and agentic in expectation. The 14 intent categories cover definition recall, follow-up clarification, exam preparation, concept explanation, code debugging, navigation, logistics, and related study behaviors, with a residual other/unmatched class.

We compared these labels against an independent LLM-based labeler over all 833
messages using the same label space (Table~\ref{tab:label_agreement}). Both
sides are automated, so the values measure consistency between two automated
methods, not accuracy against human ground truth. Per-class agreement for
interaction form, the variable Table~\ref{tab:interaction_form} rests on, is
high for questions (F1 0.972, $n = 469$) and keyword fragments (0.933,
$n = 219$), lower for imperative commands (0.817, $n = 51$), and lowest for
the residual statement/other class (0.766, $n = 94$). Intent labels are
reported only as exploratory categories, and no headline claim depends on
them.

\begin{table}[ht]
\centering
\small
\begin{tabular}{lrrr}
\toprule
Variable & Agreement & $\kappa$ & Macro-F1 \\
\midrule
Interaction form & 92.4\% & 0.876 & 0.872 \\
Refusal & 91.7\% & 0.806 & 0.903 \\
Agentic expectation & 93.5\% & 0.590 & 0.795 \\
Primary intent & 37.3\% & 0.275 & 0.303 \\
\bottomrule
\end{tabular}
\caption{Agreement between the original deterministic labels and an
independent LLM-based labeler over all 833 messages.}
\label{tab:label_agreement}
\end{table}

\subsection{Production Grounding Audit}
\label{sec:audit}

\subsubsection{Sampling and Judge}

The audit uses the four citation-by-refusal cells of Section~\ref{sec:rq2}
plus all imperative messages as an overlapping behavioral sample. It censuses
the two small cells (39 cited refusals, 44 uncited non-refusals) and all 51
imperative messages at weight 1.0, and samples 60 of the 548 cited
non-refusals (weight 9.13) and 40 of the 202 uncited refusals (weight 5.05),
giving 234 selections and 224 unique messages after deduplication, weighted
back to the 833-message population. The imperative overlay is analyzed
separately and is not a fifth population stratum. The two quotas were set by
an annotation budget rather than a power analysis; with denominators up to
118, the best-supported estimates carry roughly $\pm$10 percentage points.

Judgments come from \emph{nvidia/nemotron-3-ultra-550b-a55b} and are stored in the audit artifacts for each record.
Decoding used temperature 0.0 with a fixed seed, and of 1,107 records
attempted, 1,106 parsed successfully, with the one failure marked failed
rather than assigned an invented label. The judge belongs to a different model
family from the deployed \modelname generator, but both sides remain
automated, so the audit is diagnostic rather than human validation.

\subsubsection{Estimates and Their Limits}

Retrieval relevance asks whether cited evidence is relevant to the question,
citation support asks whether that evidence fully, partially, or does not
support the generated answer, and refusal appropriateness asks whether
declining was justified by the evidence available to the judge.
Table~\ref{tab:audit_full} reports the weighted estimates. Intervals are
stratified bootstrap percentile intervals: a Wilson interval assumes an
unweighted binomial and is not valid for a weighted stratified estimator, so
Wilson intervals are used only for the unweighted production proportions in
Table~\ref{tab:grounding_results}, such as citation coverage and the explicit
refusal rate.

\begin{table}[ht]
\centering
\small
\setlength{\tabcolsep}{3pt}
\begin{tabular}{lrrr}
\toprule
Estimate (weighted) & Value & 95\% CI & $n$ \\
\midrule
Full citation support & 0.650 & 0.550--0.750 & 80 \\
Full or partial support & 0.863 & 0.788--0.938 & 80 \\
No citation support & 0.113 & 0.050--0.188 & 80 \\
Irrelevant-evidence citations & 0.017 & 0.001--0.044 & 118 \\
Appropriate refusals & 0.619 & 0.514--0.718 & 99 \\
Implicit refusal, uncited & 0.614 & 0.477--0.750 & 44 \\
\bottomrule
\end{tabular}
\caption{Weighted diagnostic estimates. $n$ is the audited denominator for
each estimate.}
\label{tab:audit_full}
\end{table}

A separate diagnostic audit of 50 uncited messages labeled 54\% of the queries
clearly unanswerable from the available course evidence and 46\% unclear, and
rated 58\% of the refusals appropriate, 38\% unclear, and 4\% inappropriate.
The dominant labeled failure reasons were insufficient evidence to judge and
missing course content, at 40\% each.

Both audits share one limit. The judge's course-content inventory was
assembled from passages cited elsewhere in the same course, so material that
retrieval never surfaced was invisible to the judge. We therefore do not
report an answerable-but-missed population estimate, which this construction
would bias toward zero.

\subsection{Offline Retrieval Evaluation Details}
\label{sec:ablation_details}

\subsubsection{Corpus, Arms, and Configuration}
An offline benchmark was necessary because the production courses have no gold
retrieval labels: student questions in the logs carry no annotated correct
segment, so retrieval quality cannot be scored against them.

The evaluation uses EduVidQA \citep{ray2025eduvidqa}, restricted to videos
with usable transcripts and course assignments: 10 courses, 139 videos, 1,062
chapters, and 5,924 chunks. The real-world split has 269 questions, all
timestamped; the synthetic split has 1,056 questions of which only 18 carry
timestamps, so timestamp-dependent metrics are not reported for it. Evaluation
timestamps and source-video identifiers are used only for measurement and
never enter retrieval, and chapter assignments are reused across arms rather than regenerated per run.

Four arms are compared: dense-only cosine retrieval; dense plus BM25 without
the summary prior; the complete system adding the chapter-summary soft prior;
and the complete system with course isolation removed, scoring all 5,924
chunks per query. All arms use 250-word chunks with 50-word overlap,
\embeddingmodel embeddings, a seven-chunk retrieval budget, and a
12,000-character context budget.
\subsubsection{Metrics and Significance}

Timestamp hit equals 1 when a retrieved chunk from the source video covers the gold timestamp within a tolerance of $\pm 30$ seconds. Video hit equals 1 when any retrieved chunk comes from the source video; a no-evidence result counts as a miss. Precision@5 is the fraction of the five retrieval positions occupied by chunks from the source video. MRR is the reciprocal rank of the first timestamp-relevant chunk. nDCG@5 is $\mathrm{DCG}@5/\mathrm{IDCG}@5$, where
\[
\mathrm{DCG}@5 =
\sum_{i=0}^{4}
\frac{\mathrm{rel}_i}{\log_2(i+2)}
\]
and $\mathrm{rel}_i$ is binary timestamp relevance.

Binary metrics use an exact McNemar test over paired questions and ranked
metrics use the Wilcoxon signed-rank test(Table~\ref{tab:offline_significance}). Against dense-only
retrieval, the complete system produced 29 timestamp-hit wins, 226 ties, and
14 losses, and 29 wins, 228 ties, and 12 losses for video hit. Removing course
isolation is significantly worse under any possible discordant-pair
configuration, with $p \leq 0.0005$ for video hit and $p \leq 0.034$ for
timestamp hit.

\begin{table}[ht]
\centering
\small
\setlength{\tabcolsep}{3pt}
\begin{tabular}{lrrrr}
\toprule
Metric & Dense & Complete & Diff. & $p$ \\
\midrule
Timestamp hit & 0.4015 & 0.4572 & 0.0558 & 0.0315 \\
Video hit & 0.6840 & 0.7472 & 0.0632 & 0.0115 \\
MRR & 0.2504 & 0.2934 & 0.0431 & 0.0239 \\
nDCG@5 & 0.2826 & 0.3299 & 0.0473 & 0.0103 \\
\bottomrule
\end{tabular}
\caption{Paired comparison between dense-only retrieval and the complete
system over 269 real-world questions.}
\label{tab:offline_significance}
\end{table}

\subsubsection{Sensitivity, Synthetic Split, and Reproducibility}

The prior weight $\gamma$ was varied while preserving the dense-to-BM25 ratio, with $\alpha + \beta = 1 - \gamma$ (Table~\ref{tab:gamma}). Performance was strongest at $\gamma=0.1$ and generally declined as the chapter-summary prior received more weight. The narrow usable range is further evidence that the prior is a modest refinement rather than the
load-bearing component.

\begin{table}[ht]
\centering
\small
\setlength{\tabcolsep}{4pt}
\begin{tabular}{lrrrr}
\toprule
$\gamma$ & TS hit & Video hit & MRR & nDCG@5 \\
\midrule
0.1 & 0.4610 & 0.7584 & 0.2916 & 0.3291 \\
0.2 & 0.4424 & 0.7472 & 0.2909 & 0.3221 \\
0.3 & 0.4424 & 0.7398 & 0.2910 & 0.3243 \\
0.4 & 0.4349 & 0.7249 & 0.2851 & 0.3174 \\
0.5 & 0.4126 & 0.6840 & 0.2709 & 0.3026 \\
\bottomrule
\end{tabular}
\caption{Sensitivity to the chapter-summary-prior weight, holding the
dense-to-BM25 ratio fixed ($n = 269$).}
\label{tab:gamma}
\end{table}

On the synthetic split, the complete system increased video hit from 0.834 to
0.860 and video precision from 0.529 to 0.588 relative to dense-only
retrieval. Per-course results on the real-world split show the complete system
improving or matching video hit in six of seven courses, with the one decrease
on a course where the dense baseline was already strong; three of the seven
courses have fewer than 30 questions and are directional only.

The evaluation was run twice in separate processes, with the encoder
pinned to CPU and 32-bit floating point, fusion computed in 64-bit floating
point, and ties resolved deterministically by chunk identifier. Across 807 query-arm comparisons (three arms over 269 questions; the no-isolation control was not included in the stability run), the two runs produced zero mismatches in retrieved chunk identities, scores, per-query metrics, aggregate metrics, and significance inputs within a tolerance of $10^{-12}$.

\subsection{Additional Tables}

Table~\ref{tab:course} reports Course-level behavioral contrasts. Q-wds = mean query words; Ref.\ =
refusal rate; HW-paste = homework-paste rate. Small-course values are
directional only. The two remaining active courses recorded 7 and 1 messages
and are omitted as too small to interpret. HW-paste rates come from the
original deployment analysis; the detection rule was not recovered for
re-verification.

\begin{table}[H]
\centering
\scriptsize
\setlength{\tabcolsep}{2.5pt}
\begin{tabular}{lrrrrrr}
\toprule
Course & Msgs & Q-wds & Cited & Ref. & HW-paste & Agentic \\
\midrule
BIOL 2321 & 656 & 20.1 & 74.7\% & 26.2\% & 2.7\% & 7.9\% \\
BIOL 2301 & 57 & 5.9 & 43.9\% & 49.1\% & 0.0\% & 7.0\% \\
COSC 1336 & 47 & 6.6 & 68.1\% & 27.7\% & 4.3\% & 4.3\% \\
COSC 4393 & 40 & 14.6 & 72.5\% & 25.0\% & 47.5\% & 2.5\% \\
BIOL 4315/6315 & 14 & 10.6 & 50.0\% & 42.9\% & 0.0\% & 14.3\% \\
BIOL 2302 & 11 & 8.3 & 18.2\% & 63.6\% & 0.0\% & 0.0\% \\
\bottomrule
\end{tabular}
\caption{Course-level behavioral contrasts.}
\label{tab:course}
\end{table}

\end{document}